\documentclass{article}
\pdfoutput=1
\usepackage{ijcai22}

\usepackage{times}

\usepackage{soul}
\usepackage{url}
\usepackage[hidelinks]{hyperref}
\usepackage[utf8]{inputenc}
\usepackage[small]{caption}
\usepackage{graphicx}
\usepackage{amsmath}
\usepackage{booktabs}
\usepackage{algorithm}
\usepackage{algorithmic}
\usepackage{epsfig}
\usepackage{amsmath}
\usepackage{amssymb}
\usepackage{framed,multirow}
\usepackage{latexsym}
\usepackage{xcolor}
\usepackage{multirow}
\usepackage[normalem]{ulem}
\usepackage{arydshln}
\usepackage{subfig}

\newcommand{\mb}[1]{\mathbf{#1}}

\newcommand{\tb}[1]{\textbf{#1}}
\newcommand{\tu}[1]{\underline{#1}}

\title{Self-Supervised Depth Estimation with Isometric-Self-Sample-Based Learning}

\author{
Geonho Cha, Ho-Deok Jang, Dongyoon Wee\\
\affiliations
Clova AI, NAVER Corp.\\\{geonho.cha, hodeok.jang, dongyoon.wee\}@navercorp.com}

\begin{document}

\maketitle

\begin{abstract}
Managing the dynamic regions in the photometric loss formulation has been a main issue for handling the self-supervised depth estimation problem.
Most previous methods have alleviated this issue by removing the dynamic regions in the photometric loss formulation based on the masks estimated from another module, making it difficult to fully utilize the training images.
In this paper, to handle this problem, we propose an isometric self-sample-based learning (ISSL) method to fully utilize the training images in a simple yet effective way.
The proposed method provides additional supervision during training using self-generated images that comply with pure static scene assumption.
Specifically, the isometric self-sample generator synthesizes self-samples for each training image by applying random rigid transformations on the estimated depth. Thus both the generated self-samples and the corresponding training image always follow the static scene assumption.
We show that plugging our ISSL module into several existing models consistently improves the performance by a large margin.
In addition, it also boosts the depth accuracy over different types of scene, i.e., outdoor scenes (KITTI and Make3D) and indoor scene (NYUv2), validating its high effectiveness.
\end{abstract}


\section{Introduction}

Though estimating depth from a single RGB image is one of the core problems in computer vision, it has been considered an ill-posed problem due to its inherent ambiguities and expensive annotation cost.
To alleviate these difficulties, recent self-supervised learning methods \cite{zhou:unsupdepth:cvpr2017,mahjourian:icpdepth:cvpr2018,zou:dfnet:eccv18,ranjan:compeclloa:cvpr2019,godard:digging:iccv19,guizilini:3dpacking:cvpr2020,klingner:semanticguidance:eccv20,shu2020featdepth} use image sequences during the training based on the static scene assumption.
The static scene assumption presumes the deformation between consecutive images as a rigid transformation in 3D space to train a depth estimator with self-supervised losses such as the photometric loss \cite{zhou:unsupdepth:cvpr2017}.

However, it is not easy to satisfy the static scene assumption in real-world data because dynamic regions frequently occur by moving objects or occlusions.
Here, we use the term ``dynamic regions" to denote the image regions that do not satisfy the static scene assumption.
Thus, dynamic regions in training images have been considered one of the principal reasons for degrading the depth estimation performance.

To deal with dynamic regions, most of the previous studies have excluded the dynamic regions in the photometric loss calculation based on the masks estimated from another module \cite{zhou:unsupdepth:cvpr2017,ranjan:compeclloa:cvpr2019,klingner:semanticguidance:eccv20} or induced from the geometry consistency prior \cite{bian:unsupervised:neurips2019}.
However, these approaches cannot fully utilize the training images due to the excluded dynamic regions in the photometric loss formulation, which could hamper the estimation performance, especially in the dynamic regions.

In this paper, to resolve this issue, we propose an isometric-self-sample-based learning (ISSL) method, which consists of the isometric self-sample generator and the corresponding loss.
The isometric self-sample generator generates self-samples in a way that allows us to use the entire image including dynamic region during training.
To generate self-samples, the isometric self-sample generator performs rigid transformations to the point cloud estimated from an image of the training dataset.
The self-samples are generated by sampling RGB values from the training image with the correspondences determined by projecting the transformed point cloud into the image space.
Because each self-sample is associated with the corresponding training sample by a single rigid transformation in 3D space, all the self-samples generated by our approach satisfy the static scene assumption.

For modeling self-supervision on self-samples, we apply geometric consistency loss~\cite{bian:unsupervised:neurips2019} between the self-samples and the corresponding training image.
The loss is applied in conjunction with the conventional self-supervised losses, providing an additional relative depth supervision in a self-supervised manner.
Our modular design enables ISSL to be utilized as a universal plug-and-play module to boost the performance.

The proposed method can effectively improve the depth estimation performance in case that the baseline depth estimator infers depths with little shape error.
Under this condition, the proposed method improves the translation error with the relative depth self-supervision, resulting in an improved depth inference performance.
Please refer to Section \ref{sec:analysis} for the detailed discussion.

In experiments, we show that our approach consistently improves several existing models by large margin. In addition, it boosts the depth accuracy over different types of scenes, i.e., outdoor scenes (KITTI \cite{Geiger:kitti:cvpr2012} and Make3D \cite{saxena:make3d:tpami2008}) and indoor scene (NYUv2 \cite{Silberman:ECCV12}), validating its high effectiveness.
The contributions of our method are summarized as below:
\begin{itemize}
    \item We propose a novel ISSL module which utilizes self-samples which are generated in a way that meet the static scene assumption with the corresponding training image. 
    \item The proposed scheme consistently boosts several existing monocular depth estimation models by a large margin. It also shows high effectiveness on different types of scenes such as KITTI and Make3D (outdoor scenes) and NYUv2 (indoor scene) benchmark datasets.
\end{itemize}

\begin{figure*}[t]
    \centering
    \includegraphics[height=5.5cm]{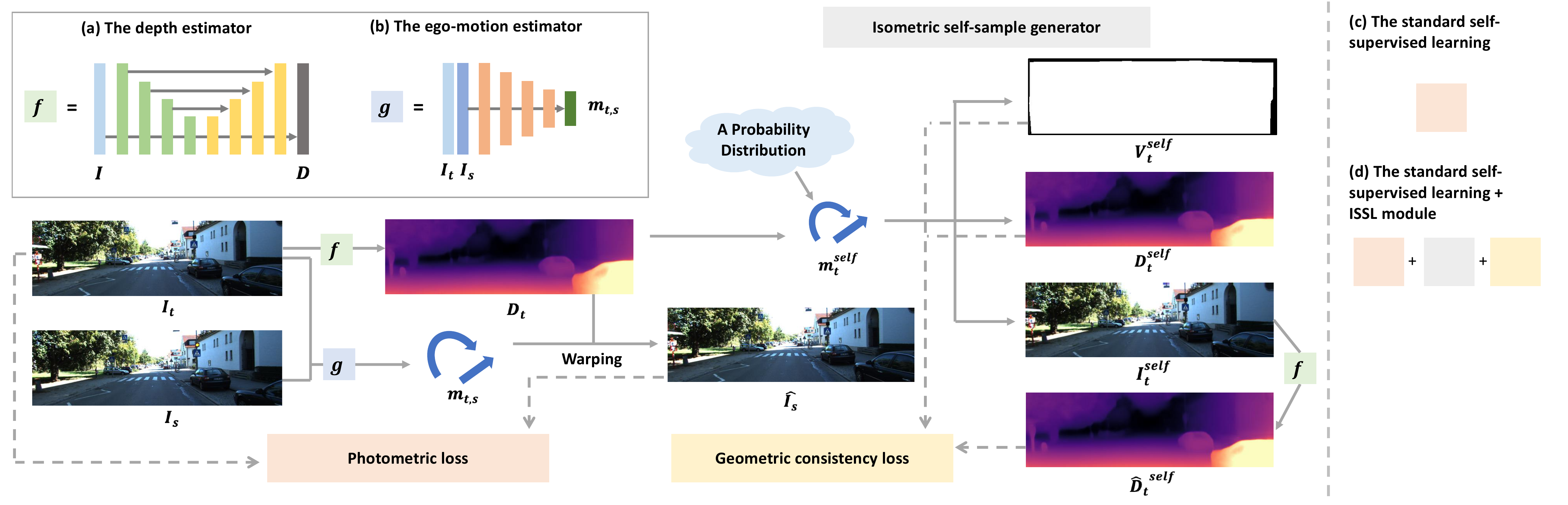}
    \caption{\textbf{An overview of ISSL.} We omitted the edge-aware smoothness loss for simplicity. (a) The depth estimator infers the depth map for a given single RGB image. (b) The ego-motion estimator infers the camera ego-motion for a given pair of RGB images. (c, d) Unlike the standard self-supervised learning methods which train the depth and ego-motion estimator only based on the photometric loss, our method additionally utilizes isometric consistency in 3D space via isometric self-samples. The isometric self-sample generator generates self-samples in a way that the self-samples follow the strict static scene assumption over the entire image regions. To this end, the isometric self-sample generator utilizes point cloud which is obtained based on the estimated depth map. Some rigid transformations are applied to the point cloud to synthesize self-samples.}
    \label{fig:overview}
\end{figure*}

\section{Related work}
After \cite{zhou:unsupdepth:cvpr2017} proposed the self-supervised monocular depth estimation framework, which only exploits monocular videos for the training, plenty of methods have been proposed to improve the performance.
Since the self-supervised photometric loss is formulated based on the static scene assumption which is frequently dissatisfied due to moving objects or occlusions, handling the dynamic regions in training images has been a key for the performance improvement.

One of the main approaches to handling the dynamic regions is to filter out the dynamic regions utilizing information from the other tasks.
\cite{zou:dfnet:eccv18} proposed the method which estimates depth and optical flow at the same time.
They proposed the cross-task consistency loss which measures the discrepancy between the flows inferred from the two tasks.
Here, the dynamic regions are detected based on the forward-backward consistency of the estimated flows.
\cite{ranjan:compeclloa:cvpr2019} tackled the multi-task problem of estimating monocular depth, optical flow, and motion segmentation.
They formulated the problem as a game: the depth estimator and the optical flow estimator compete to model the input images, and the motion segmentation estimator moderates the two players.
The motion segmentation estimator detects the dynamic regions which are filtered out in the photometric loss calculation.
\cite{gordon:depthitw:iccv2019} addressed the multi-task problem of estimating depth, camera ego-motion, and camera intrinsics.
They handled the dynamic region problem based on the considerations of the geometric depth relation.
\cite{chen:selfcamera:iccv2019} also addressed the multi-task problem of estimating depth, camera ego-motion, optical flow, and camera intrinsics.
They handled the dynamic region based on the min-pooling operation between the ego-motion flow and the optical flow.
\cite{klingner:semanticguidance:eccv20} addressed the multi-task learning problem of depth estimation and semantic segmentation.
From the estimated semantic information, the image regions of the moving object categories were filtered out in formulating the loss function.
On the one hand, \cite{bian:unsupervised:neurips2019} filtered out the dynamic region in a different way.
They proposed the geometry consistency loss which measures the discrepancy of estimated target and source depth maps.
They showed that the geometry consistency can be used to infer masks of dynamic regions.

There have been several methods which utilize the full training image region.
\cite{dai2020self} utilized full training image regions by incorporating the estimated six degree-of-freedom object motion in calculating the photometric loss.
However, they need segmentation masks estimated from another big module \cite{he2017mask} which is trained with expensive supervision.
\cite{li2020unsupervised} incorporated full image regions in the training by inferring object motion map.
To handle the ill-posedness of the problem, they proposed some regularization losses following the prior knowledge of the motion map.
However, they only modeled the translation part of the object motion, which can degrade the performance when the objects have rotative motions.

It is to be noted that most of the previous work have handled the problem of dynamic region by filtering out the dynamic regions when formulating the photometric loss, which inhibits the training image utilization.
In order to address this problem, we propose ISSL, which is a simple yet effective method for the performance improvement by fully utilizing the training images.

Meanwhile, several methods were proposed to increase the depth estimation performance by improving network structure or proposing learning techniques.
\cite{zhou:dualnetwork:iccv19} proposed a dual networks architecture, which can effectively handle high-resolution images as input and efficiently generate high-resolution depth maps.
\cite{godard:digging:iccv19} proposed various techniques such as per-pixel minimum reprojection, auto-masking stationary pixels, and multi-scale estimation that improve the performance of the depth estimator.
\cite{guizilini:3dpacking:cvpr2020} proposed PackNet that leverages symmetrical packing and unpacking blocks which help to preserve detailed information from an RGB image.

Note here that our method can improve the existing models in an orthogonal manner, implying its high efficacy as a universal technique to boost the performance.

\section{Method}
An overview of ISSL is visualized in Figure \ref{fig:overview}.
We first explain the standard setting of the self-supervised monocular depth estimation and then present the proposed method.

\subsection{Self-supervised monocular depth estimation}
\label{sec:unsu_monodepth}
The self-supervised depth estimation aims to train a depth estimator $f(\cdot)$, which estimates the depth map $\mb{D}$, of an RGB image $\mb{I}$, without any depth supervision.
To train without the depth supervision, consecutive images from monocular videos are used during the training to formulate self-supervised loss functions.
When designing the loss functions, we assume that the deformations between neighboring images can be modeled with the rigid ego-motion estimated from the ego-motion estimator $g(\cdot)$.
A key component of the self-supervised losses is the photometric loss which measures the discrepancy of the RGB value distributions between the target image and the target-view-synthesized neighboring source images.

To explain the warping process, let us consider a target image $\mb{I}_t$, and the corresponding neighboring source images $\mb{I}_s^i$, $i=1,\cdots,n_s$, in a training image sequence, where $n_s$ is the number of the source images considered.
To carry out the view-synthesis from a source image to the target image, we first estimate the depth map of the target image and ego-motion from the target image to each source image as
\begin{equation}
    \mb{D}_t = f(\mb{I}_t), ~~\mb{m}_{t, s}^i=g(\mb{I}_t, \mb{I}_s^i),
\end{equation}
where $\mb{D}_t$ is the estimated depth map of $\mb{I}_t$, and $\mb{m}_{t, s}^i$ is the estimated ego-motion from the target image to the $i$-th source image which is the six-dimensional vector consisting of three-dimensional rotation values in the form of axis-angle representation and three-dimensional translation.

With the estimated depth map and the ego-motions, we can warp each source image to the target image based on the geometric relation between the target image and each source image in 3D space \cite{mahjourian:icpdepth:cvpr2018}.
In the following explanation, we assume homogeneous coordinates.
First, we lift the estimated depth map of the target image to the point cloud in 3D space as
\begin{equation}
    \mb{p}_{t}^{uv} = D_{t}^{uv}\mb{K}^{-1}[u,v,1]^T,
\end{equation}
where $u$ and $v$ are the image coordinates, $\mb{K}$ is the given camera intrinsic matrix, $\mb{p}_{t}^{uv}$ and $D_{t}^{uv}$ are the 3D point and the depth value corresponds to the $(u, v)$-th pixel of the target image, respectively.
After that, we transform the point cloud of the target image to the view of each source image with the estimated ego-motion as
\begin{equation}
    \mb{\hat{p}}_{s}^{i, uv} = \mb{M}_{t, s}^i\mb{p}_{t}^{uv},
\end{equation}
where $\mb{M}_{t, s}^i$ is the transformation matrix corresponds to $\mb{m}_{t,s}^i$.
By projecting the $\mb{\hat{p}}_{s}^i$ onto the image plane, the pixel-wise correspondences between the target image and each source image can be obtained, so as we can get the target-view-synthesized image $\mb{\hat{I}}_{s}^i$, of each source image.
Since the sampling pixels may not have integer coordinates, the values are computed by bilinear interpolation of values of its four closest pixels.
Based on the synthesized images, the photometric loss is defined as
\begin{equation}
    \begin{split}
    L_{\textrm{p}} = \sum_{t}\sum_{i} \alpha L_{\textrm{1}}(\mb{\hat{I}}_{s}^i, \mb{I}_t)+(1-\alpha)L_{\textrm{SSIM}}(\mb{\hat{I}}_{s}^i, \mb{I}_t),
    \end{split}
\end{equation}
where $L_{\textrm{1}}(\cdot)$ is the $L_1$ loss, $L_{\textrm{SSIM}}(\cdot)$ is the structured similarity loss \cite{wang:ssim:tip2004}, and $\alpha$ is a weighting parameter.

Along with the photometric loss, we incorporate the edge-aware smoothness prior loss to regularize the depth in texture-less image regions, following the practices in \cite{godard:monodepth:cvpr2017,godard:digging:iccv19,guizilini:3dpacking:cvpr2020}.
The edge-aware smoothness prior loss is defined as
\begin{equation}
    L_{\textrm{s}} = \sum_t |\partial_x \mb{\tilde{D}}_t|e^{-|\partial_x \mb{I}_t|}+|\partial_y \mb{\tilde{D}}_t|e^{-|\partial_y \mb{I}_t|},
\end{equation}
where $|\cdot|$ denotes the absolute value, $\mb{\tilde{D}}_t$ is the mean-normalized inverse depth map of the target image, $\partial_x$ and $\partial_y$ are the partial gradient along $x$-axis and $y$-axis, respectively.

It is to be noted that the photometric loss is formulated based on the assumption that the deformation from the source image to the target image can be modeled with a rigid transformation.
However, it can be hard to meet in general situations because non-rigid deformations between the pair of images frequently occur by moving objects or occlusions.
Most of the previous frameworks have handled this issue by filtering out dynamic regions from training images when designing the photometric loss.
However, in this way, we cannot fully utilize the training images, which could degrade the performance of the trained depth estimator.
To handle this issue, we propose ISSL, which is explained in the next section.

\subsection{Isometric-self-sample-based learning}
\label{sec:proposed}
We first explain the details of the isometric self-sample generation procedure, then introduce the loss function which is applied for the generated self-samples. Lastly, we present the total loss for training the proposed framework.\\\\
\textbf{Isometric self-sample generation}\\
For a target image $\mb{I}_t$, we first obtain the point cloud of the target image $\mb{p}_t$, based on the depth map $\mb{D}_{t}$, inferred from the depth estimator.
After that, we apply rigid transforms $\mb{m}_{t, k}^{\textrm{self}}$, $k=1,\cdots,n_k$, to the point cloud, where $n_k$ is the number of generated self-samples for each target image.
Here, the rigid transform parameters are randomly sampled from a probability distribution:
\begin{equation}
    \mb{m}_{t, k}^{\textrm{self}} \sim \mathcal{P}(\theta_p),
\end{equation}
where $\mathcal{P}$ denotes a probability distribution, and $\theta_p$ is the parameter for the distribution.

We obtain the $k$-th self-sample $\mb{I}_{t, k}^{\textrm{self}}$, by sampling the RGB values from the target image based on the correspondences determined by projecting the transformed 3D point cloud.
In a similar way, we obtain $\mb{D}_{t, k}^{\textrm{self}}$, which is used along with $\mb{I}_{t, k}^{\textrm{self}}$ in the formulation of the loss function, by sampling the depth values from $\mb{D}_{t}$.
Additionally, we calculate the validity mask $\mb{V}_{t, k}$, of which the value is one if the 3D point is projected onto the proper image region and zero otherwise.
In the self-sample generation process, the sampling pixels may not have integer coordinates, similar to the synthesizing process of $\mb{\hat{I}}_{s}$.
Hence, the values are computed by bilinear interpolation of values of its four closest pixels. 
Note here that the proposed self-sample generation is completely different from 2D image augmentations such as random scaling and rotation in that the transforms for the self-sample generation are carried out in 3D space.
Some visualizations of the generated self-samples can be found in Appendix.\\\\
\textbf{Loss function}\\
For modeling self-supervision of generated self-samples, we apply geometric consistency~\cite{bian:unsupervised:neurips2019} loss to measure the discrepancy between the estimated depth map of the $k$-th self sample $\mb{\hat{D}}_{t, k}^{\textrm{self}}$, and its corresponding depth map $\mb{D}_{t, k}^{\textrm{self}}$.
Note here that we have to resolve the scale ambiguity, which is inherent in the monocular depth estimation problem, when we compare two depth maps.
We address this issue by performing median scaling before measuring the discrepancy (An ablative result of the median scaling can be found in Appendix).
The loss is defined as
\begin{equation}
    L_{\textrm{issl}} = \sum_{t}\sum_{k}\sum_{u,v}\mb{V}_{t, k}^{uv}\cdot\frac{|\mb{\hat{D}}_{t, k}^{\textrm{self}, uv}-\mb{D}_{t, k}^{\textrm{self}, uv}|}{\mb{\hat{D}}_{t, k}^{\textrm{self}, uv}+\mb{D}_{t, k}^{\textrm{self}, uv}}.
\end{equation}
\textbf{Total loss}\\
The proposed method is trained based on the following loss
\begin{equation}
    L = \lambda_1 L_p + \lambda_2 L_s + \lambda_3 L_{issl},
\end{equation}
where $\lambda_1$, $\lambda_2$, and $\lambda_3$ are weighting parameters.
Here, the total loss is averaged over batch, self-samples, and pixels.
Please refer to Appendix for detailed implementation details.


\begin{table*}
\caption{Performance comparison results on the KITTI dataset with the ground truth \protect\cite{eigen:multiscale:neurips2014}. For the type, ``M'' means the self-supervised monocular supervision, and ``S'' means the self-supervised stereo supervision. In case of the dataset, ``K'' means that the network is trained on the KITTI dataset, and ``CS+K'' means the network is pre-trained on the cityscape dataset and fine-tuned to the KITTI dataset. In each category, the best results are in bold, and the second best are underlined.}
\centering
\resizebox{0.85\textwidth}{!}{%
\begin{tabular}{l|c|cc|cccc|ccc} \toprule
\multirow{2}{*}{Method} & \multirow{2}{*}{Type} & \multirow{2}{*}{Resolution} &\multirow{2}{*}{Dataset} & \multicolumn{4}{|c|}{Error $\downarrow$} & \multicolumn{3}{|c}{Accuracy $\uparrow$} \\
\cline{5-11} & & & & AbsRel & SqRel & RMS & RMSlog & $< 1.25$ & $<1.25^2$ & $<1.25^3$ \\ \toprule
Struct2Depth \cite{casser:structure2depth:aaai2019} & M & 416x128 & K & 0.141 & 1.026 & 5.291 & 0.215 & 0.816 & 0.945 & 0.979 \\
SGDepth \cite{klingner:semanticguidance:eccv20} & M & 640x192 & K & 0.117 & 0.907 & 4.844 & 0.196 & 0.875 & 0.958 & 0.980 \\
Monodepth2 \cite{godard:digging:iccv19}& M & 640x192 & K & 0.115 & 0.903 & 4.863 & 0.193 & \tb{0.877} & \tu{0.959} & \tu{0.981} \\
PackNet-SfM \cite{guizilini:3dpacking:cvpr2020} & M & 640x192 & K & \tu{0.113} & \tu{0.840} & \tu{4.659} & \tu{0.191} & \tb{0.877} & \tu{0.959} & \tu{0.981} \\
\textbf{Monodepth2+ISSL (Ours)} & M & 640x192 & K & \tb{0.110} & \tb{0.763} & \tb{4.648} & \tb{0.184} & \tb{0.877} & \tb{0.961} & \tb{0.983} \\

\hline
SfMLearner \cite{zhou:unsupdepth:cvpr2017} & M & 416x128 & CS+K & 0.198 & 1.836 & 6.565 & 0.275 & 0.718 & 0.901 & 0.960 \\
Vid2Depth \cite{mahjourian:icpdepth:cvpr2018} & M & 416x128 & CS+K & 0.159 & 1.231 & 5,912 & 0.243 & 0.784 & 0.923 & 0.970 \\
DF-Net \cite{zou:dfnet:eccv18} & M & 576x160 & CS+K & \tu{0.146} & \tu{1.182} & \tu{5.215} & \tu{0.213} & \tu{0.818} & \tu{0.943} & \tu{0.978} \\
\textbf{Monodepth2+ISSL (Ours)} & M & 640x192 & CS+K & \tb{0.106} & \tb{0.749} & \tb{4.546} & \tb{0.181} & \tb{0.885} & \tb{0.963} & \tb{0.983} \\
\hline

Zhou et al. \cite{zhou:dualnetwork:iccv19} & M & 1248x384 & K & 0.121 & \tu{0.837} & 4.945 & 0.197 & 0.853 & 0.955 & \tu{0.982} \\
Monodepth2 \cite{godard:digging:iccv19} & M & 1024x320 & K & 0.115 & 0.882 & 4.701 & \tu{0.190} & \tu{0.879} & \tu{0.961} & \tu{0.982} \\
SGDepth \cite{klingner:semanticguidance:eccv20} & M & 1280x384 & K & \tu{0.113} & 0.880 & \tu{4.695} & 0.192 & \tb{0.884} & \tu{0.961} & 0.981 \\
\textbf{Monodepth2+ISSL (Ours)} & M & 1280x384 & K & \tb{0.112} & \tb{0.703} & \tb{4.495} & \tb{0.185} & 0.878 & \tb{0.963} & \tb{0.984} \\
\hline
Garg \cite{garg:geometrytorescue:eccv16}& S & 640x192 & K & 0.152 & 1.226 & 5.849 & 0.246 & 0.784 & 0.921 & 0.967 \\
Monodepth R50 \cite{godard:monodepth:cvpr2017}& S & 1280x384 & K & 0.133 & 1.142 & 5.533 & 0.230 & 0.830 & 0.936 & 0.970 \\
StrAT \cite{mehta2018structured} & S & 640x192 & K & 0.128 & 1.019 & 5.403 & 0.227 & 0.827 & 0.935 & 0.971 \\
3Net (R50) \cite{poggi2018learning}& S & 640x192 & K & 0.129 & 0.996 & 5.281 & 0.223 & 0.831 & 0.939 & 0.974 \\
3Net (VGG) \cite{poggi2018learning}& S & 640x192 & K & 0.119 & 1.201 & 5.888 & 0.208 & 0.844 & 0.941 & \tb{0.978} \\
SuperDepth+pp (R50) \cite{pillai2019superdepth}& S & 1280x384 & K & 0.112 & 0.875 & \tu{4.958} & \tu{0.207} & 0.852 & 0.947 & \tu{0.977} \\
Monodepth2 \cite{godard:digging:iccv19}& S & 640x192 & K & \tu{0.109} & \tu{0.873} & 4.960 & 0.209 & \tu{0.864} & \tu{0.948} & 0.975 \\
\textbf{Monodepth2+ISSL (Ours)} & S & 640x192 & K & \tb{0.106} & \tb{0.817} & \tb{4.838} & \tb{0.199} & \tb{0.871} & \tb{0.953} & \tb{0.978} \\
\hline
FeatDepth \cite{shu2020featdepth} & MS & 1024x320 & K & \tu{0.099} & \tb{0.697} & \tu{4.427} & \tu{0.184} & \tu{0.889} & \tu{0.963} & \tu{0.982} \\

\textbf{FeatDepth+ISSL (Ours)} & MS & 1024x320 & K & \tb{0.096} & \tu{0.700} & \tb{4.334} & \tb{0.179} & \tb{0.894} & \tb{0.964} & \tb{0.983} \\
\bottomrule
\end{tabular}%
}
\label{tab:kitti_original}
\end{table*}

\section{Experiments}
We present experimental results on the depth estimation on challenging KITTI \cite{Geiger:kitti:cvpr2012}, Make3D \cite{saxena:make3d:tpami2008}, and NYUv2 \cite{Silberman:ECCV12} benchmark datasets. For training and evaluation protocol, we follow generally used ones from \cite{eigen:multiscale:neurips2014}, \cite{godard:digging:iccv19}, and \cite{bian2021unsupervised} for each benchmark.
For the quantitative evaluation, we incorporate the two types of generally used measures \cite{godard:digging:iccv19,guizilini:3dpacking:cvpr2020}: error-based and accuracy-based measures.
Note here that we could not reason the scale parameter in the monocular depth estimation setting.
To handle this, we perform the median scaling before the evaluation.

\subsection{KITTI benchmark}

We compare the results of our models with the ones from recent models, over several types of self-supervision: monocular video (M), stereo image pairs (S). The results are summarized in Table \ref{tab:kitti_original}.
Note that the performance reported in \cite{guizilini:3dpacking:cvpr2020} is the post-processed performance, which can be found in Appendix.
The results show that our model improves several baseline methods \cite{godard:digging:iccv19,shu2020featdepth} by a large margin, validating the high efficacy of the proposed method. In addition, our models are also highly benefited from a larger amount of training data as seen in pre-training experiments (CS+K), showing its high scalability. We also show additional results for analyzing the impact of using higher resolutions (1280×384). It indicates that the high resolution provides additional boosts, especially over large error cases (e.g., SqRel and RMS). 
Lastly, our approach consistently benefits the models from various input types such as monocular videos (M) and stereo image pairs (S).
We also report the post-processed quantitative comparison results with other recent methods in Appendix.

\begin{figure}[t]
    \centering
    \includegraphics[width=1.0\linewidth]{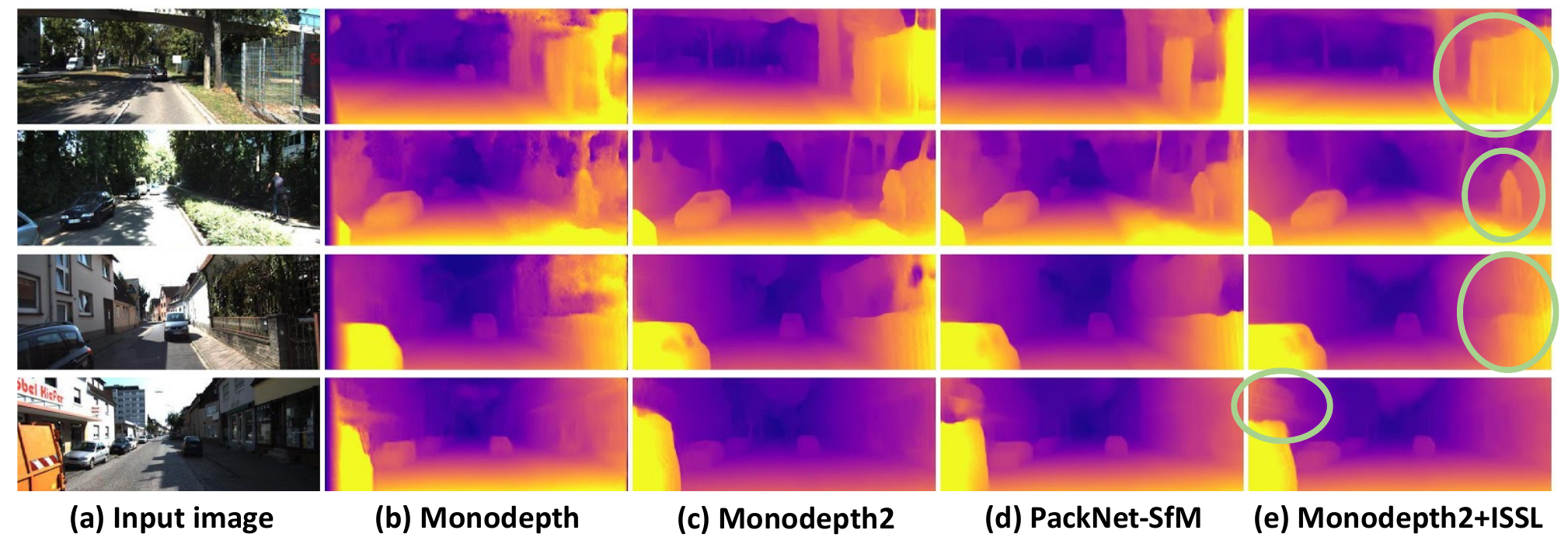}
    \caption{Qualitative results on the KITTI dataset. Our method shows more accurate results compared to the other methods.}
    \label{fig:kitti_merged}
\end{figure}

We provide a qualitative comparison with other recent models over diverse scenes in Figure \ref{fig:kitti_merged}.
We can see that the proposed method shows better or comparable estimation qualities over the other models in all the cases.
Specifically, in the first case, the other methods fail to estimate the depth of the fence.
In the second case, the proposed method successfully estimated the depth of a person, which is failed in other methods.
In the remaining cases, the proposed method successfully reconstructed the depth in the region highlighted by the green circles, which is failed in other methods.

\begin{table}
\centering
\caption{Performance comparison results on the Make3D dataset. ``D'' means the network is trained with the depth supervision.}
\resizebox{0.45\textwidth}{!}{%
\begin{tabular}{l|c|cccc} \toprule
\multirow{2}{*}{Method} & \multirow{2}{*}{Type} & \multicolumn{4}{|c}{Error $\downarrow$}\\
\cline{3-6} & & AbsRel & SqRel & RMS & $\textrm{log}_{\textrm{10}}$ \\
\toprule
Karsch \cite{karsch:depthtransfer:tpami2014}& D & \tu{0.428} & \tu{5.079} & \tu{8.389} & \tu{0.149} \\
Liu \cite{liu:dcde:cvpr2014}& D & 0.475 & 6.562 & 10.05 & 0.165 \\
Laina \cite{laina:ddpw:3dv16}& D & \tb{0.204} & \tb{1.840} & \tb{5.683} & \tb{0.084} \\
\hline
Zhou \cite{zhou:unsupdepth:cvpr2017}& M & 0.383 & 5.321 & 10.470 & 0.478 \\
DDVO \cite{wang:ddvo:cvpr2018}& M & 0.387 & 4.720 & 8.090 & 0.204 \\
PackNet-SfM \cite{guizilini:3dpacking:cvpr2020}& M & 0.378 & 4.367 & 8.221 & 0.184 \\
Monodepth2 \cite{godard:digging:iccv19}& M & \tu{0.322} & \tu{3.589} & \tu{7.417} & \tu{0.163} \\
\textbf{Monodepth2+ISSL (Ours)} & M & \tb{0.316} & \tb{2.938} & \tb{6.863} & \tb{0.160} \\
\bottomrule
\end{tabular}%
}

\label{tab:make3d}
\end{table}

\begin{table}
\caption{NYUv2 evaluation results. The experiment is conducted on the self-supervised monocular supervision setting.}
\centering
\resizebox{0.45\textwidth}{!}{%
\begin{tabular}{l|cccc|ccc} \toprule
\multirow{2}{*}{Method} & \multicolumn{4}{|c|}{Error $\downarrow$} & \multicolumn{3}{|c}{Accuracy $\uparrow$} \\
\cline{2-8} & AbsRel & SqRel & RMS & RMSlog & $< 1.25$ & $<1.25^2$ & $<1.25^3$ \\ \toprule
Monodepth2 & \tu{0.196} & \tu{0.216} & \tu{0.745} & \tu{0.367} & \tu{0.732} & \tu{0.895} & \tu{0.940} \\
Ours & \tb{0.156} & \tb{0.123} & \tb{0.561} & \tb{0.211} & \tb{0.784} & \tb{0.939} & \tb{0.979} \\
\bottomrule
\end{tabular}%
}
\label{tab:nyuv2}
\end{table}

\subsection{Make3D and NYUv2 benchmark}
We report the performance on the Make3D dataset using models trained on KITTI. Following the evaluation practices from \cite{godard:digging:iccv19}, we use image that is center cropped with 2x1 ratio. The input resolution is set to 512x256, and the depth is capped to $70$m.
The results are summarized in Table \ref{tab:make3d}.
We can see that our model outperforms all methods trained using self-supervision by a large margin, which demonstrates the generalization performance of our model.

We also evaluate the performance of our method on NYUv2 \cite{Silberman:ECCV12} dataset. For the training, we use the rectified images of \cite{bian2021unsupervised}. The results are summarized in Table \ref{tab:nyuv2}. We can demonstrate that the proposed method is effective not only on the outdoor scenes (i.e., KITTI and Make3D) but also on the indoor scene. Here, the unsatisfactory performance of \cite{godard:digging:iccv19} comes from the training collapse issue reported in \cite{bian2021unsupervised}. However, despite this issue, we can see that the proposed method improves the performance by a large margin.

\section{Analysis}
\label{sec:analysis}
\subsection{Experimental setup}
In order to identify the impact of the additional depth supervision from ISSL module, we analyze the improvement of depth error caused by ISSL during training.
We define dynamic region as moving instance area and static region as the residual area. 
Specifically, we extract the moving instance area using both the semantic segmentation results inferred from the pretrained model of \cite{klingner:semanticguidance:eccv20} and the instance segmentation results inferred from the pretrained MaskRCNN of Pytorch library \cite{paszke2019pytorch}. In addition, we decompose the depth error from dynamic region into two parts: shape error and translation error. Here, shape error is measured by adjusting the estimated depth by median of ground-truth depth in an instance-wise manner. Then, translation error is acquired as residual one. All results are evaluated on KITTI benchmark.

\subsection{Results}
\begin{figure}[t]%
\centering
{{\includegraphics[width=0.88\linewidth]{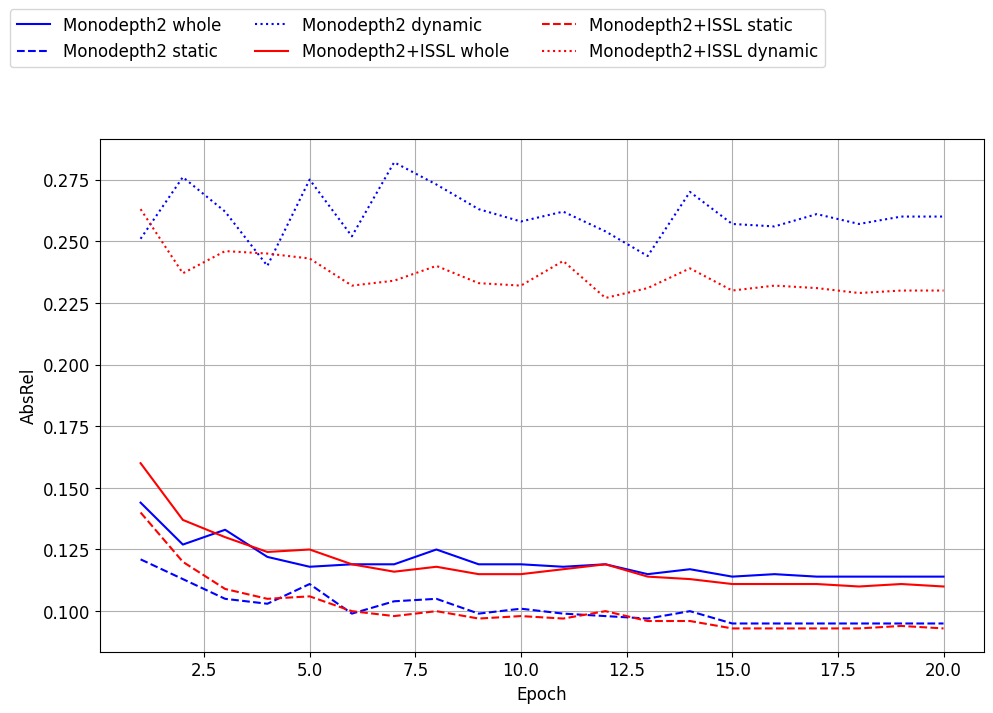} }}%
\hfill
{{\includegraphics[width=0.88\linewidth]{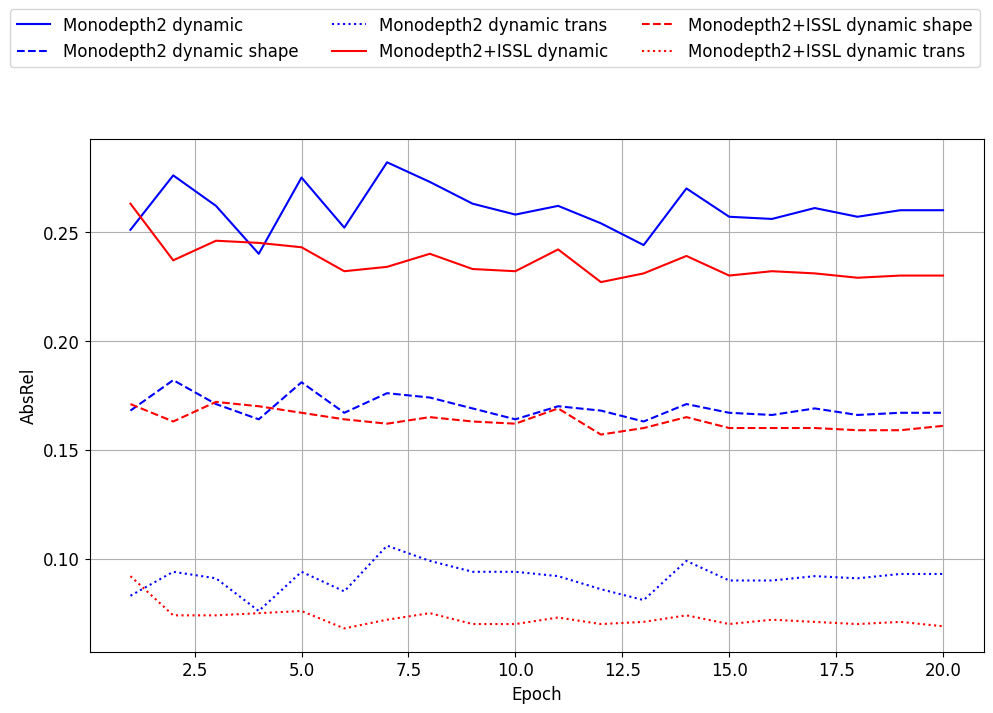} }}%
\caption{Impact of ISSL module on depth error during training over KITTI benchmark. It improves dynamic depth error by a large margin along with slight gain in static depth error, contributing to the overall gain in whole depth error. Here, trans. denotes translation.}%
\label{fig:analysis}
\end{figure}

First, we analyze the impact of our approach on the depth error from static and dynamic region. As shown in Figure \ref{fig:analysis} (top), we can clearly see that our method highly improves depth error in dynamic region along with slight gain in static region, both contributing to the overall gain in whole depth error. In addition, we provide analysis on the depth error from dynamic region of the baseline model~\cite{godard:digging:iccv19} by disentangling it into shape and translation error. From Figure~\ref{fig:analysis} (bottom), we can clearly see that translation error takes non-trivial portion in dynamic depth error. It is mainly originated from moving object which breaks the static scene assumption and induces the miss of pixel correspondence under photometric self-supervision. Thus, the translation error is not improved during training and shape error is not minimized accordingly. On the other hand, the self-sample generated by our ISSL module is associated with the corresponding training sample by a single rigid transformation in 3D space. Thus, self-sample becomes free from translation error in dynamic region, enabling compliance with the static scene assumption. Although there can be some noise from shape error, more effective supervision is available compared to the baseline. When we train with our self-sample, the translation error is greatly improved and the shape error is also slightly benefited, leading to the large overall improvement of dynamic depth error. Table \ref{tab:kitti_abl_dynamic} shows the numbers of Figure \ref{fig:analysis} (bottom) at the final epoch.

\begin{table}
\caption{Impact of ISSL module on dynamic depth error. Our module improves both shape and trans. error, especially helping trans. error by a large margin. Both gains contribute to the overall boost of the dynamic depth error. Here, trans. denotes translation.}
\centering
\resizebox{0.43\textwidth}{!}{%
\begin{tabular}{c|l|cccc} \toprule
\multirow{2}{*}{Error type} & \multirow{2}{*}{Model} & \multicolumn{4}{c}{Dynamic region} \\
\cline{3-6}
 & & AbsRel & SqRel & RMS   & RMSlog\\
\hline
\multirow{2}{*}{whole}  & Monodepth2      & 0.260  & 4.468 & 9.280 & 0.296 \\
                        & Monodepth2+ISSL & 0.230  & 3.119 & 8.548 & 0.278 \\
\hline
\multirow{2}{*}{shape}  & Monodepth2      & 0.167  & 2.405 & 6.921 & 0.281 \\
                        & Monodepth2+ISSL & 0.161  & 1.921 & 6.500 & 0.249 \\
\hline
\multirow{2}{*}{trans.} & Monodepth2      & 0.093  & 2.063 & 2.359 & 0.015 \\
                        & Monodepth2+ISSL & 0.069  & 1.198 & 2.048 & 0.029 \\
\bottomrule
\end{tabular}%
}

\label{tab:kitti_abl_dynamic}
\end{table}

\section{Conclusion}
\label{sec:conclusion}
We have proposed the isometric-self-sample-based learning method.
It enables full utilization of the training images, which has been difficult in the previous works.
The experimental results demonstrate the large performance boosts on several benchmark datasets, showing its effectiveness.
Note that we could design another consistency loss in the domain of the ego-motions, which provides additional supervision to the ego-motion estimator.
It could be beneficial for the performance improvements, which is left as a future work.

\bibliographystyle{named}
\bibliography{ref}

\setcounter{section}{0}
\renewcommand\thesection{\Alph{section}}
\renewcommand\thesubsection{\thesection.\arabic{subsection}}

\clearpage

\section{Appendix}
\subsection{Dataset}
\label{sec:dataset}
\noindent
\textbf{KITTI dataset.} The KITTI dataset is widely used for the performance evaluation of the depth estimation methods.
It consists of image sequences captured from the camera attached to the driving car.
This dataset has ground-truth depth information collected using LiDAR sensors.
We use the data split of \cite{eigen:multiscale:neurips2014} for the fair comparison.
As a result, the training, validation, and test set consists of $39810$, $4424$, $697$ images, respectively.\\\\
\textbf{Cityscape dataset.} The cityscape dataset consists of image sequences captured on the urban street.
This dataset has no ground-truth depth information for the quantitative evaluation.
Hence, the cityscape dataset is used for pretraining the network, which is finetuned to the KITTI dataset.
We use the same data preprocessing protocol from \cite{zhou:unsupdepth:cvpr2017,godard:monodepth:cvpr2017}: cropping out the bottom 25\% of the image to remove the car logo and subsampling every two frames to match the frame rate of the KITTI dataset.
As a result, it consists of $88250$ training images.\\\\
\textbf{Make3D dataset.} The Make3D dataset consists of $400$ training images and $134$ test images with the aligned depth map data.
Since the number of training images is too small for the network training, this dataset is generally used to validate the generalization performance. We first train the proposed framework on the KITTI dataset, and evaluate the depth estimation performance on the test images of the Make3D dataset.

\subsection{Implementation details}
For the depth and ego-motion estimator, we adopt the same models from \cite{godard:digging:iccv19} with ResNet18 \cite{kaiming:resnet:cvpr2016} as encoder which is pre-trained on ImageNet \cite{olga:imagenet:ijcv15}. We also follow the main objective function and all the techniques introduced in \cite{godard:digging:iccv19} such as per-pixel minimum reprojection error and auto-masking. Please refer to \cite{godard:digging:iccv19} for more detailed model configurations. To feed a set of source images for training monocular models, we use preceding and succeeding frames, i.e., $n_s=2$.

For the isometric self-sample generator, we sample rigid transformation parameter from a uniform distribution. The sampling range is parameterized by $\theta_p$, which consists of $\theta_R$ and $\theta_T$.
Each denotes the maximum absolute range of uniform distribution for sampling rotation and translation parameters, respectively. During training, we use $\theta_R$ linearly increased from $0.005$ to $0.2$ every epoch, and constant $\theta_T$ of $0.005$. We generate four self-samples for each target image (i.e., $n_k=4$).
Here, artifacts, holes, and collisions in the self-samples could be issues, which could occur by inaccurately estimated depth or large rigid deformations.
To mitigate these issues, we set the sampling parameters with moderate values which are linearly increased every epoch to allow the self-sample generator to use more precise depths when incorporating relatively large rigid deformations.

To build our framework, we use PyTorch \cite{paszke2019pytorch}. For optimization, we use Adam \cite{kingma2017adam} optimizer with initial learning rate of $1\times 10^{-4}$, which is decreased by a factor of $10$ after $15$ epochs. The whole training ends at $20$ epochs. In the case of pre-training on Cityscapes \cite{cordts:cityscape:cvpr2016}, all models are trained for $10$ epochs using the same initial learning rate without any scheduling. The batch size is set to $12$.
For balancing each term of the objective function, we use $\lambda_1=1$, $\lambda_2=0.001$, $\lambda_3=0.1$, and $\alpha = 0.15$, unless stated otherwise. Our high resolution models are trained for five epochs with the learning rate of $1\times 10^{-5}$ starting from the pretrained weights for $10$ epochs with the standard resolution (640x192) setting. The batch size is set to three in this case. For applying our framework to \cite{shu2020featdepth}, we use the same parameter setting used in \cite{shu2020featdepth} and above, except for $\lambda_3=0.01$.


\subsection{Additional experiments}
\begin{figure}[t]
    \centering
    \includegraphics[width=0.99\columnwidth]{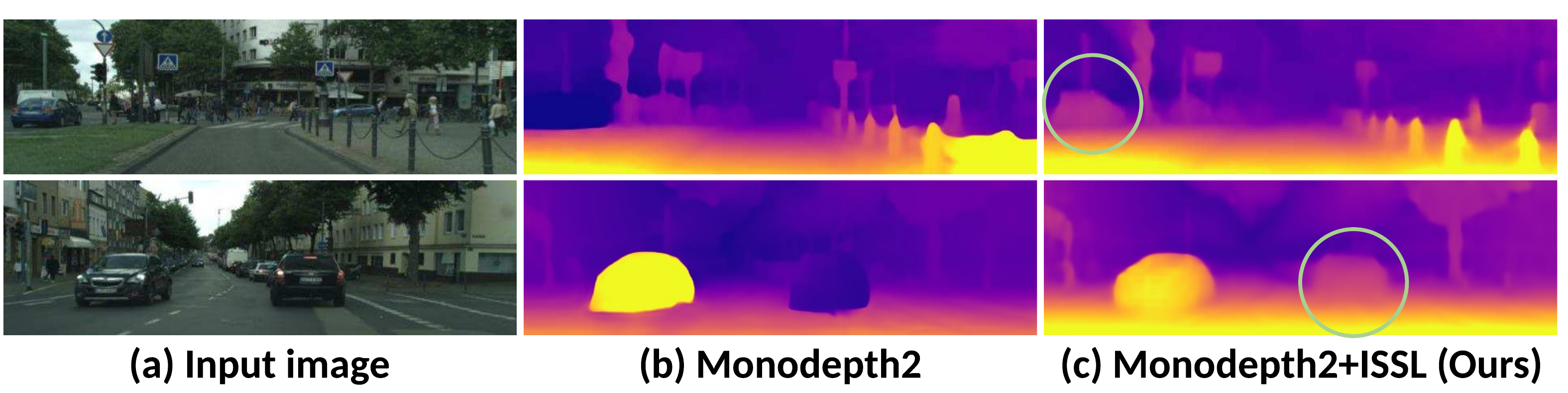}
    \caption{Qualitative comparison results on Cityscape dataset.}
    \label{fig:qual_cityscape}
\end{figure}

\noindent
\textbf{KITTI results with post-processing.}
We report the post-processed KITTI benchmark comparison results with other recent methods \cite{godard:digging:iccv19,guizilini:3dpacking:cvpr2020} in Table \ref{tab:kitti_original_pp}.
For the post processing, we adopt the method proposed in \cite{godard:monodepth:cvpr2017}.
In all the cases, the proposed method shows better or competitive performances compared to the other methods, demonstrating the effectiveness of our method.\\\\
\textbf{Effectiveness of ISSL on cityscapes dataset.}
We report a qualitative result on the cityscape dataset. For the evaluation, we split the cityscape dataset into disjoint train and test sets. The qualitative results on the test set are visualized in Figure \ref{fig:qual_cityscape}. The figure shows that the proposed method is effective to rectify holes of dynamic objects shown in baseline.\\\\
\textbf{Ablation study.} To further validate the proposed approach, we conduct ablation studies on the KITTI dataset.
The results are summarized in Table \ref{tab:kitti_ablation_a}.
We try the normal distribution for sampling rigid transformation parameters.
Here, the standard deviation is set to the half of $\theta_P$ used in the uniform distribution, and the sampled parameters are clipped to the maximum absolute value becomes $\theta_p$ used in the uniform distribution setting. This setting results in similar performance as uniform distribution, showing its robustness over the choice of random distribution (Exp 2, 4).
On the other hand, without median scaling in the calculation of the isometric consistency loss, the performance is slightly degraded (Exp 3, 4).
It also results in reduced scale-consistency, increasing the standard deviation of scaling parameters in test set from $2.137$ to $2.429$. 


\begin{figure*}
    \centering
    \includegraphics[width=1.90\columnwidth]{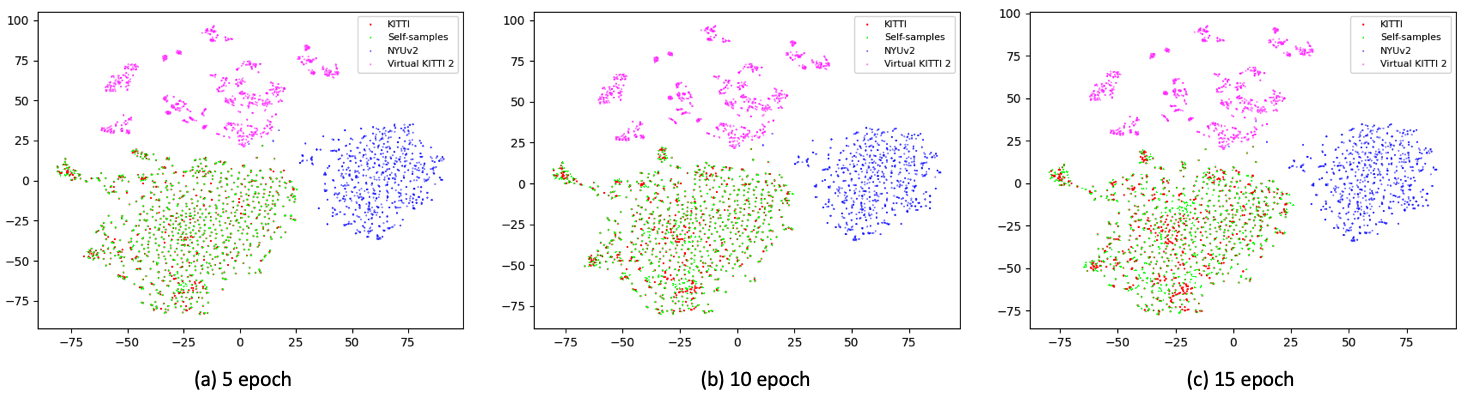}
    \caption{A t-SNE visualization of the LPIPS features of the generated self-samples.}
    \label{fig:ftsne}
\end{figure*}

\begin{table*}
\caption{Post-processed performance comparison results on the KITTI dataset with the ground truth \protect\cite{eigen:multiscale:neurips2014}. We used the post-processing scheme introduced in \protect\cite{godard:monodepth:cvpr2017}. For the type, ``M'' means the self-supervised monocular supervision. In case of the dataset, ``K'' means that the network is trained on the KITTI dataset, and ``CS+K'' means the network is pre-trained on the cityscape dataset and fine-tuned to the KITTI dataset. In each category, the best results are in bold, and the second best are underlined.}
\centering
\resizebox{0.97\textwidth}{!}{%
\begin{tabular}{l|c|cc|cccc|ccc} \toprule
\multirow{2}{*}{Method} & \multirow{2}{*}{Type} & \multirow{2}{*}{Resolution} &\multirow{2}{*}{Dataset} & \multicolumn{4}{|c|}{Error $\downarrow$} & \multicolumn{3}{|c}{Accuracy $\uparrow$} \\
\cline{5-11} & & & & AbsRel & SqRel & RMS & RMSlog & $< 1.25$ & $<1.25^2$ & $<1.25^3$ \\ \toprule
Monodepth2+pp \cite{godard:digging:iccv19}& M & 640x192 & K & 0.112 & 0.851 & 4.754 & 0.190 & \tb{0.881} & \tu{0.960} & 0.981 \\
PackNet-SfM+pp \cite{guizilini:3dpacking:cvpr2020} & M & 640x192 & K & \tu{0.111} & \tu{0.785} & \tb{4.601} & \tu{0.189} & \tu{0.878} & \tu{0.960} & \tu{0.982} \\
\textbf{Monodepth2+ISSL+pp} & M & 640x192 & K & \tb{0.110} & \tb{0.754} & \tu{4.621} & \tb{0.183} & \tu{0.878} & \tb{0.961} & \tb{0.983} \\

\hline
PackNet-SfM+pp \cite{guizilini:3dpacking:cvpr2020} & M & 640x192 & CS+K & \tu{0.108} & \tb{0.727} & \tb{4.426} & \tu{0.184} & \tu{0.885} & \tu{0.963} & \tb{0.983} \\
\textbf{Monodepth2+ISSL+pp} & M & 640x192 & CS+K & \tb{0.105} & \tu{0.732} & \tu{4.503} & \tb{0.180} & \tb{0.887} & \tb{0.964} & \tb{0.983} \\
\hline

Monodepth2+pp \cite{godard:digging:iccv19} & M & 1024x320 & K & 0.112 & 0.838 & 4.607 & 0.187 & \tu{0.883} & \tu{0.962} & \tu{0.982} \\
PackNet-SfM+pp \cite{guizilini:3dpacking:cvpr2020} & M & 1280x384 & K & \tb{0.107} & \tu{0.802} & \tu{4.538} & \tu{0.186} & \tb{0.889} & \tu{0.962} & 0.981 \\
\textbf{Monodepth2+ISSL+pp} & M & 1280x384 & K & \tu{0.111} & \tb{0.693} & \tb{4.473} & \tb{0.184} & 0.878 & \tb{0.963} & \tb{0.984} \\
\bottomrule
\end{tabular}%
}
\label{tab:kitti_original_pp}
\end{table*}

\begin{table*}
\caption{Ablation results on the KITTI dataset with the standard setting, i.e., no pretraining on cityscape, 640x192 resolution, no post processing. Here, Monodepth2 is used as the baseline.}
\centering
\resizebox{0.85\textwidth}{!}{%
\begin{tabular}{c|l|cccc|ccc} \toprule
\multirow{2}{*}{Exp} & \multirow{2}{*}{Method} & \multicolumn{4}{|c|}{Error $\downarrow$} & \multicolumn{3}{|c}{Accuracy $\uparrow$} \\
\cline{3-9} & & AbsRel & SqRel & RMS & RMSlog & $< 1.25$ & $<1.25^2$ & $<1.25^3$ \\ \toprule
1 (baseline) & Ours w/o $L_{\textrm{issg}}$  & 0.115 & 0.916 & 4.857 & 0.192 & 0.878 & 0.960 & 0.981 \\
2 & Ours w/ Gaussian & 0.111 & 0.787 & 4.663 & 0.186 & 0.878 & 0.960 & 0.982 \\
3 & Ours w/o median scaling & 0.111 & 0.789 & 4.660 & 0.185 & 0.878 & 0.961 & 0.983 \\
\hline
4 & Ours (full) & 0.110 & 0.763 & 4.648 & 0.184 & 0.877 & 0.961 & 0.983 \\
\bottomrule
\end{tabular}%
}
\label{tab:kitti_ablation_a}
\end{table*}

\begin{figure*}
    \centering
    \includegraphics[width=1.85\columnwidth]{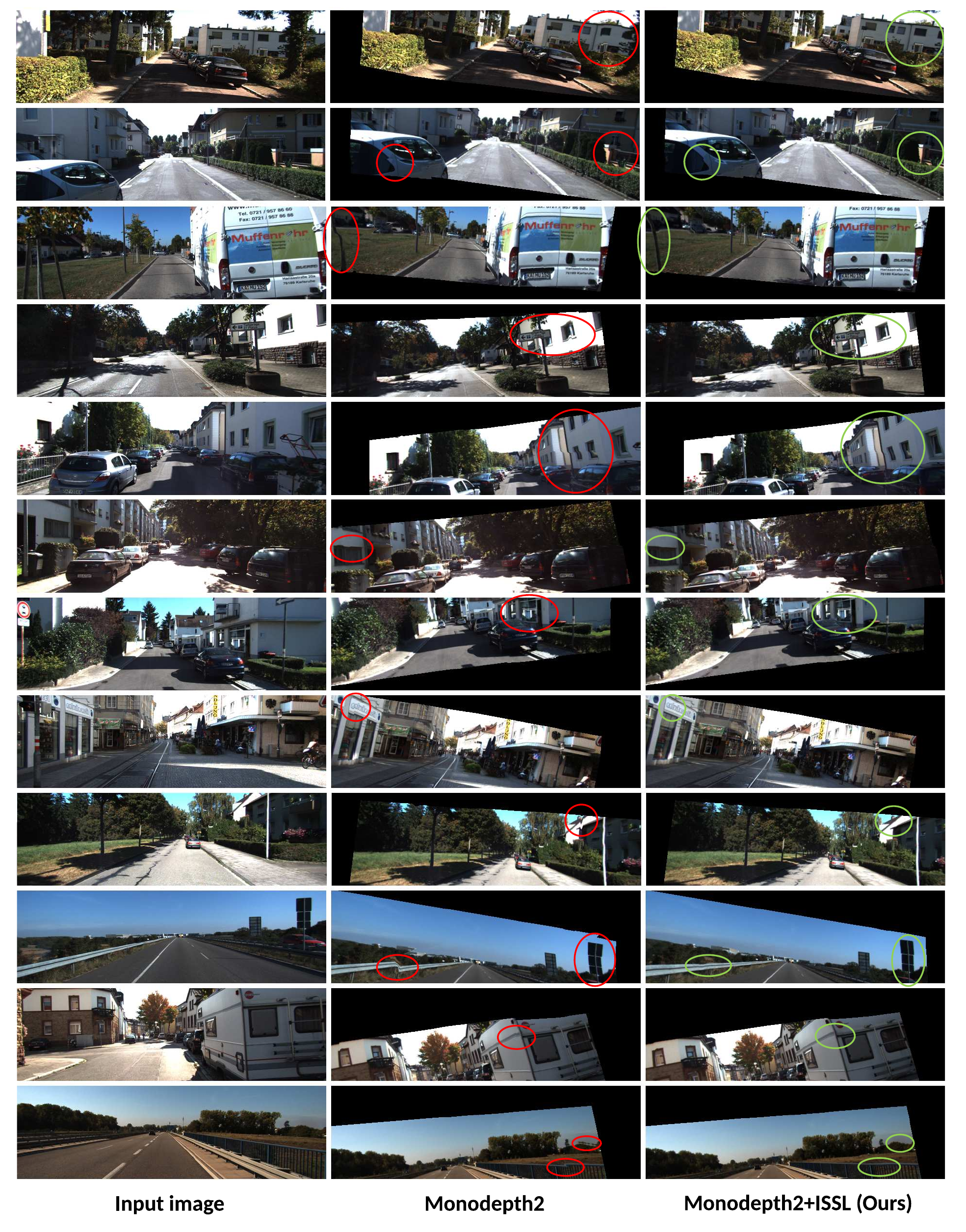}
    \caption{A visualization of the generated self-samples. (From left to right) Each image represents an input image, the generated self-sample from the baseline, and the generated self-sample from the proposed framework, respectively. Here, the same rigid transformation parameters are applied to generate the self-samples in both the cases.}
    \label{fig:gen_self_samples_3}
\end{figure*}

\subsection{Visualization of the generated self-samples}
\noindent
\textbf{Visualization in feature space.} The performance could be degraded if the quality of the self-samples is not satisfactory or the distribution of the self-samples is too different from that of the training images.
To check if this is the case, we compare the LPIPS features \cite{zhang2018perceptual} of the self-samples to various image sets: KITTI training images, NYUv2 images, and Virtual KITTI 2 images \cite{cabon2020vkitti2}.
A t-SNE plot is visualized in Figure \ref{fig:ftsne}.
From the figure, we can verify that the LPIPS features of self-samples are closer to those of KITTI training images compared to those of other datasets like simulated Virtual KITT2. Accordingly, we speculate that self-samples can provide effective augmentations with stochastic transformations in the valid distribution.\\\\
\textbf{Visualization in RGB space.} To verify the effect of the proposed isometric consistency loss to the generated self-samples, we visualize two types of self-samples which are generated based on two methods: (i) the depth map is estimated from the baseline, and (ii) the depth map is estimated from the proposed framework.
The generated self-samples are illustrated in Figure \ref{fig:gen_self_samples_3}.
Here, the visualized self-samples are generated in the training process, and the same rigid transformation parameters are applied in both the cases for the fair comparison.
We can see that the use of the proposed isometric consistency loss tends to reduce the artifacts when we compare the two types of self-samples.
Note that the artifacts would not appear in the generated self-samples if the depth map is estimated with less error. Here, we do not consider the issues come from self-occlusion and disocclusion.
Therefore, we can consider the existence of the artifacts as a qualitative measure.
In this sense, we can demonstrate that the proposed framework performs better compared to the baseline.

\end{document}